\documentclass{article}

\usepackage{microtype}
\usepackage{graphicx}
\usepackage{subcaption}
\usepackage{booktabs} 

\usepackage{hyperref}

\usepackage[accepted]{icml2026}
\usepackage{multirow}
\usepackage{amsmath}
\usepackage{amssymb}
\usepackage{mathtools}
\usepackage{amsthm}
\usepackage{marvosym}
\usepackage{xcolor}
\usepackage[table]{xcolor}
\usepackage[most]{tcolorbox}
\newtcolorbox{calloutbox}[2][]{%
  enhanced,                 
  breakable,                
  colframe=black,           
  boxrule=1.1pt,            
  arc=2mm,                  
  left=6mm,right=6mm,       
  top=3mm,bottom=3mm,       
  colbacktitle=black,       
  coltitle=white,           
  colback=gray!6,           
  fonttitle=\bfseries,      
  title={#2},               
  #1                        
}

\newtcolorbox{simplebox}[1][]{%
  enhanced,                 
  breakable,                
  colframe=black,           
  boxrule=1pt,              
  arc=0.5mm,                  
  left=2mm,right=1mm,       
  top=2mm,bottom=1mm,       
  colback=cyan!2,           
  #1                        
}

\definecolor{myblue}{RGB}{30,144,255}

\usepackage[capitalize,noabbrev]{cleveref}

\theoremstyle{plain}

\theoremstyle{definition}

\theoremstyle{remark}

\usepackage[textsize=tiny]{todonotes}

\icmltitlerunning{Deep Residual Injection for Full-Spectrum Forensic Signal Perception in Multimodal Large Language Models}

\begin{document}

\twocolumn[
  \icmltitle{Deep Residual Injection for Full-Spectrum Forensic Signal Perception in Multimodal Large Language Models}



  \icmlsetsymbol{equal}{*}
  \icmlsetsymbol{leader}{\textdagger}
  \icmlsetsymbol{corresponding}{\Letter}

  \begin{icmlauthorlist}
    \icmlauthor{Kaiqing Lin}{equal,yyy,youtu}
    \icmlauthor{Zhiyuan Yan}{equal,comp}
    \icmlauthor{Ruoxin Chen}{equal,youtu}
    \icmlauthor{Ke-Yue Zhang }{leader,youtu}
    \icmlauthor{Yue Zhou}{yyy}
    \icmlauthor{Caiyong Piao}{fudan}
    \icmlauthor{Bin Li}{corresponding,yyy}
    \icmlauthor{Taiping Yao}{youtu}
    \icmlauthor{Bo Wang}{youtu}
    \icmlauthor{Youchang Xiao}{youtu}
    \icmlauthor{Shouhong Ding}{youtu}
  \end{icmlauthorlist}

  \icmlaffiliation{yyy}{Guangdong Provincial Key Laboratory of Intelligent Information Processing, Shenzhen Key Laboratory of Media Security, and SZU-AFS Joint Innovation Center for AI Technology, Shenzhen University}
  \icmlaffiliation{youtu}{Tencent Youtu Lab}
  \icmlaffiliation{comp}{Peking University}
  \icmlaffiliation{fudan}{Fudan University}


  \icmlcorrespondingauthor{Bin Li}{libin@szu.edu.cn}

  \icmlkeywords{AI-generated Image Detection, Multimodal Large Language Model, ICML}

  \vskip 0.3in
]



\printAffiliationsAndNotice{\icmlEqualContribution\textsuperscript{\textdagger}Project leader \textsuperscript{\Letter}Corresponding author.}

\begin{abstract}
Multimodal large language models (MLLMs) have been increasingly adopted in forensics for their robust semantic understanding.
As AI-generated images become realistic, semantic-level inconsistencies alone are often insufficient for reliable detection.
This motivates a critical question: \textit{whether MLLMs can achieve full-spectrum forensic signal perception, i.e., capturing low-level generator artifacts without sacrificing pre-trained semantic knowledge.}
We further perform a layer-wise analysis of forensic signal perception in MLLMs, showing that semantic information is primarily formed in the early-to-middle layers, whereas direct fine-tuning for artifact learning disrupts these semantic representations.
Based on this insight, we propose Deep Visual Residual MLLM (Deep-VRM) to \textit{preserve early semantic processing while injecting artifact-specific visual signals as a residual path into an intermediate layer}, where they are fused with semantic token representations and propagated through subsequent trainable layers.
This enables later layers to jointly model semantic reasoning and signal-level forensic cues, and surprisingly, the model learns to adaptively leverage different levels of forensic signals depending on the input, achieving robust and generalizable detection performance.
Extensive experiments show that our method achieves state-of-the-art across most benchmarks. The code and data are available at \url{https://github.com/KQL11/Deep-VRM}.
\end{abstract}

\begin{figure}
  \centering
  \includegraphics[width=\linewidth]{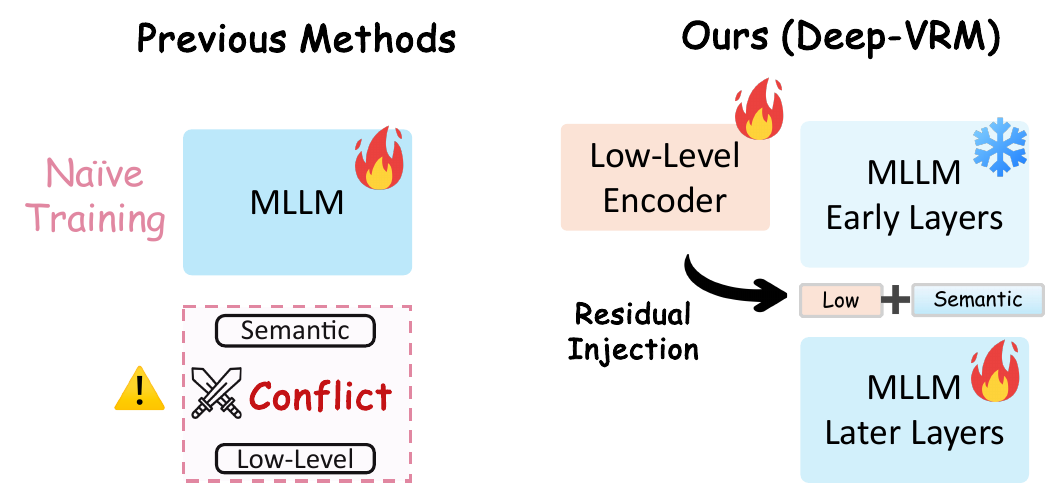}
  \caption{Comparison between naively training MLLMs for detection and ours. While previous methods suffer from a representation conflict between semantic and low-level features, our approach decouples these two spaces via residual injection. It preserves established semantic knowledge while introducing artifact sensitivity to resolve the learning dilemma.}
  \label{fig:fig1}
\end{figure}

\section{Introduction}
\label{sec:intro}
The growth of AI-generated content has precipitated a crisis in digital trust, raising urgent concerns regarding image authenticity. 
Numerous studies~\cite{legion, FakeReasoning, SpotFake, lin2025guard, he2025vlforgery} have proposed leveraging Multimodal Large Language Models (MLLMs) to detect AI-generated images and provide human-aware textual explanations. 
While MLLMs appear to be ideal candidates for detection due to their reasoning capabilities and interpretability, they face significant challenges in practice.

Generic MLLMs frequently struggle with the detection task itself, often underperforming compared to specialized forensic models~\citep{jia2024can}. 
Consequently, current solutions often resort to external expert modules (i.e. specialized forensic models) to compensate~\citep{chen2024x2,zhou2025aigi}. 
Yet, this reliance reduces the MLLM to a mere proxy rather than an independent evaluator, preventing it from learning the intrinsic features of forgery. 
Moreover, such approaches fail to elucidate the underlying why there is bad performance of the MLLMs.

In this paper, our research reveals a critical trade-off in MLLM representation learning: \textbf{native models cannot learn generalizable generator traces without compromising their core semantic capabilities, which leads to the above deficiency.}
Pretrained MLLMs are optimized for semantic features, such as image style and content and logic. 
Consequently, they inherently overlook low-level artifacts, specifically the subtle traces left by the generator, which are essential for forgery detection.
We utilize two datasets, $D_1$ (focusing on semantic features) and $D_2$ (focusing on generator traces), to explore this mechanism. 
As shown in Figure~\ref{fig:motivation}, we find that standard LoRA tuning on the original MLLM architecture struggles to recover these low-level artifacts.
Although updating more parameters (e.g., full finetuning) allows the model to learn these features, it damages its understanding of semantics (Table ~\ref{tab:commonbench}).
Consequently, achieving the dual capability required for robust detection remains a significant challenge for current architectures.

To resolve this dilemma, we should understand how the model handles these features. 
Through a layer-wise analysis using linear probes, we observe that the separability of semantic features for discriminating real from fake images is primarily established and converges in the early-to-middle layers (see Figure ~\ref{fig:motivation3}).
\textbf{This suggests that these layers are critical for extracting semantic features, while forcing these layers to learn low-level artifacts interferes with extracting semantic features in AIGI detection. }
We observe that previous MLLM-based AIGI detection methods typically process low-level artifacts and semantic features in the same manner without distinction, resulting in suboptimal performance.
Therefore, a strategy is to decouple the learning process: keeping the early layers frozen for semantics and injecting low-level artifacts in the later layers.

Guided by these insights and recognizing that the rich semantic priors of MLLMs provide a powerful foundation for detection, we propose the Deep Visual Residual MLLM (Deep-VRM), \textit{focusing on enhancing the detection performance of MLLMs, distinct from the aspect of explainability.}
As shown in Figure~\ref{fig:fig1}, our method employs a Residual Injection strategy to bypass the early layers where semantic features are encoded. 
We utilize a dual-branch visual architecture. The frozen encoder ($\mathcal{V}_o$) preserves the pretrained semantic perception already available in the MLLM, while the adapted branch ($\mathcal{V}_a$), equipped with LoRA adapters, learns artifact-specific visual cues. 
Unlike traditional methods that align features at the input level, Deep-VRM injects these artifact features directly into the deep layers of the LLM.
Our method achieves state-of-the-art performance based on an MLLM only, avoiding the need for any external expert detectors. 
By learning low-level clues after the semantic features have converged, our model ensures robust performance in wild datasets.
\begin{itemize} 
  \item We systematically analyze the MLLMs in AIGI detection and reveal a critical \textit{representation conflict}: forcing early-to-middle layers to learn low-level artifacts compromises their inherent semantic capabilities. This finding establishes that effective detection requires decoupling artifact learning from semantic preservation.

  \item We propose \textbf{Deep Visual Residual MLLM (Deep-VRM)}, a novel architecture employing a \textit{Residual Injection} strategy. 
  Unlike previous methods, Deep-VRM injects artifact-specific features directly into the deep layers of the LLM, bypassing the semantic-dominant early layers to achieve optimal feature integration.

  \item Without external detectors, Deep-VRM achieves SOTA performance solely based on the MLLM. Extensive experiments demonstrate its superior generalization in-the-wild settings, effectively capturing both subtle generator traces and high-level semantic anomalies.
\end{itemize}

\begin{figure*}
    \centering
    \includegraphics[width=\textwidth]{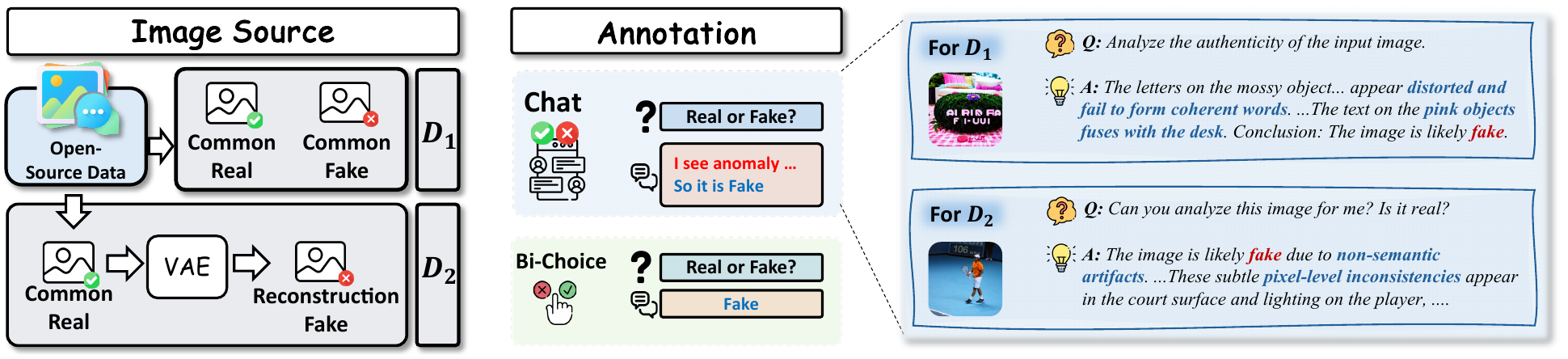}
    \caption{Illustration of the data construction pipeline. We first collect common real and fake images from various open-source datasets to form $D_1$. Subsequently, we generate reconstructed fake images via VAEs and pair them with their original real counterparts to construct $D_2$. To train the MLLM, we design two types of annotations—Chat and Bi-Choice—for all images in $D_1$ and $D_2$. Notably, for the VAE-reconstructed images in $D_2$, our annotations explicitly emphasize terms like ``non-semantic artifacts'' to guide the model in learning pixel-level discrepancies.}
    \label{fig:dataset}
\end{figure*}

\section{Related Work and Motivation}
\paragraph{Traditional AIGI Detection Methods}
For AI-generated image (AIGI) detection, initial work CNNSpot~\citep{wang2020cnn} employed standard CNNs and conducted naive training. 
While effective in detecting samples from known generators, these classifiers often struggle to generalize to unseen ones.
To address this, UnivFD~\citep{ojha2023towards} utilizes CLIP as a backbone, leveraging the robust representations of pretrained vision-language models to enhance cross-generator generalization. 
Subsequent research~\citep{liu2024forgery,tan2025c2p,zheng2024breaking,yan2025orthogonal, tan2024rethinking, yan2025dgs, yan2025ns,shen2025dino,repdfd,liu2026mirror,liu2025beyond} has further refined this by exploring advanced architectures and preprocessing techniques. 
Some efforts~\citep{tan2024frequency,chu2025fire,li2025improving,karageorgiou2025any,yan2026dual} focus on frequency-domain artifacts, identifying them as discriminative features for synthetic content. 
Moreover, recent works have explored pretraining-based strategies to further improve AIGI detection performance~\cite{zhou2026simplicity,li2025ditl2}, as well as calibration-based methods to enhance prediction reliability~\cite{li2026texture,guillaro2026quality}.
Despite these advancements, conventional detectors remain highly susceptible to image post-processing. 
In particular, their performance tends to degrade significantly when facing compressed images, raising concerns regarding their robustness in real-world scenarios.

\paragraph{MLLM-based AIGI Detection Methods}
Multimodal large language models (MLLMs) have demonstrated strong capabilities across downstream tasks, motivating their use for AIGI detection via instruction fine-tuning (IFT)~\citep{legion, FakeReasoning, SpotFake, lin2025guard, he2025vlforgery} and, more recently, reinforcement learning with self-exploration and verifiable rewards~\cite{xu2025avatarshield,xu2026genshield,nguyen2025prpo}.
By using image-text instructions, these methods guide MLLMs to identify semantic inconsistencies or visual errors.
However, MLLM-based detectors still struggle with generated images containing subtle low-level artifacts, such as generator traces.
To mitigate this limitation, some frameworks integrate external expert detectors~\citep{chen2024x2, zhou2025aigi, peng2025mllm}, while others use MLLMs as agents to coordinate multiple experts~\cite{zhu2026evoguard,yu2026agentfox}.
Although these designs introduce external forensic knowledge, they may cause the MLLM to imitate expert predictions or rely on expert coordination, rather than developing autonomous feature-level visual analysis.
Ultimately, prior research leaves two fundamental challenges unresolved: (1) how to enhance the perception of low-level forensic artifacts within the MLLM, and (2) how to balance the learning of low-level artifacts and semantic artifacts within the MLLM.
In this work, we challenge the conventional end-to-end design of MLLM-based detection by considering the intrinsic functional heterogeneity of MLLM internal layers.

\vspace*{-2mm}

\paragraph{Exploration of Functional Stratification in MLLMs}
For large language models (LLMs), prior work~\cite{skean2025layer} shows that early-layer representations are particularly sensitive to noise, whereas intermediate layers play a key role in processing complex semantic information by filtering redundant patterns and preserving discriminative features.
In the context of MLLMs, recent studies~\cite{jiang2025devils, hartman2025skip} further identify intermediate layers as a critical locus for cross-modal interaction and visual reasoning.
Together, these findings suggest a functional stratification inside LLMs and MLLMs: early-to-middle layers are crucial for forming stable semantic representations, while later layers are more responsible for high-level reasoning and information integration.
However, how such layer-wise functional specialization affects forensic signal perception remains underexplored, particularly for low-level generator artifacts in AIGI detection.
This motivates our analysis of how semantic cues and low-level artifacts are represented across MLLM layers.

\vspace*{-2mm}

\section{Method}
\label{sec:method}

To address the model's limited perception of low-level artifacts, we create a training dataset consisting of two distinct subsets.
{The first subset, $D_1$, comprises diverse images aggregated from existing training sets~\cite{russakovsky2015imagenet,unsplash_data_2025,zhu2023genimage,ye2025echo} to capture common semantic cues for detection. 
Following the data construction protocol of DDA~\cite{chen2025dual}, the second subset, $D_2$, pairs original real images from MS-COCO~\cite{coco} with their VAE-reconstructed counterparts generated using the Stable Diffusion 2.1 VAE. } 
This approach eliminates semantic interference, forcing the model to focus exclusively on generative traces inherent in the reconstruction process.
Both $D_1$ and $D_2$ are further formatted for instruction tuning. 
As illustrated in Figure~\ref{fig:dataset}, we construct two types of instruction annotations, applying both to datasets $D_1$ and $D_2$. 
The first type follows a binary-choice format (bi-choice), consisting of a query such as \textbf{\textit{``Is this image real or fake? Please answer with only a single word: `real' or `fake'.''}} and a target response such as \textbf{\textit{``fake''}}.
For the second category, we utilized Gemini 2.5 Pro to generate detailed analytical annotations regarding image content and forgery traces.
We follow the caption procedure proposed in the Forensic-Chat~\cite{lin2025seeing}.
Notably, we also annotate the fake image in $D_2$ with some words like `Non-semantic artifacts'.
By providing the model with ground-truth labels as priors, \textit{we ensured that the generated reasoning was both factually accurate}. 
Detailed statistics are provided in Table~\ref{tab:dataset}.
In total, we obtained 88,000 instruction-tuning samples.

\begin{table}[t]
\centering
\caption{Source composition of our datasets $D_{1}$ and $D_{2}$.}
\label{tab:dataset}
\resizebox{\linewidth}{!}{
\begin{tabular}{l | ccc | cccc}
\toprule
\multirow{2}{*}{\textbf{Dataset}} & \multicolumn{3}{c|}{\textbf{Real Images}} & \multicolumn{4}{c}{\textbf{Fake Images}} \\
\cmidrule(lr){2-4} \cmidrule(lr){5-8}
 & ImageNet & Ms-COCO & Unsplash & SD1.4\textsuperscript{\dag} & Echo 4o & Flux\textsuperscript{*} & VAE \\
\midrule
$D_{1}$ & 8500 & 0 & 5000 & 1000  & 250 & 3250 & 0 \\ 
$D_{2}$ & 0 & 35000 & 0 & 0 & 0 & 0 & 35000 \\ 
\bottomrule
\end{tabular}
}
\vspace{1mm}
\\
\raggedright \scriptsize{\emph{\textsuperscript{\dag} sourced from Training Set of GenImage.} \emph{\textsuperscript{*} generated by Flux-dev.}}

\end{table}

\subsection{Analysis}
\label{analysis}

\begin{table}[t]
  \centering
  \tiny
  \caption{Impact of fine-tuning strategies on general multimodal benchmarks. Evaluated via VLMEvalKit~\cite{duan2024vlmevalkit} and Ms-Swift~\cite{zhao2025swift}, the results highlight a severe performance degradation in fully fine-tuned models, indicating catastrophic forgetting.}
  \setlength{\tabcolsep}{10pt}
  \label{tab:commonbench}
  \begin{tabular}{lccc}
    \toprule
    \textbf{Model} &\textbf{BLINK} &\textbf{RealWorldVQA} &\textbf{MME} \\
    \midrule
    Backbone &0.5481 &0.6758 &1677 \\
    full finetuning LLM on $D_2$ &0.0373 &0.1137 &506 \\
    \midrule
    Ours on $D_2$ &0.5476 &0.6721 &1636 \\
    \bottomrule
  \end{tabular}
\end{table}

\begin{figure}[t]
  \centering
  \includegraphics[width=\linewidth]{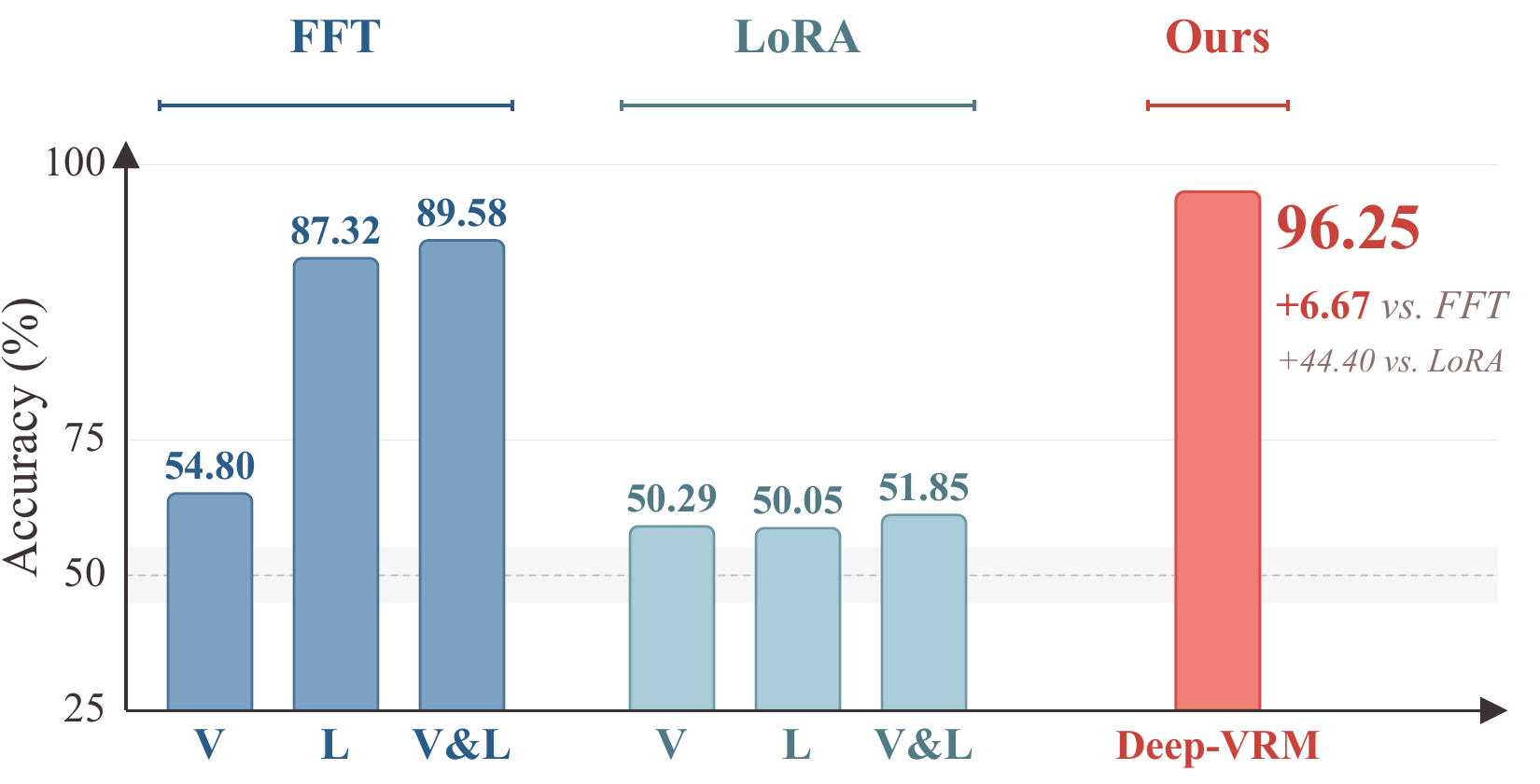}
  \caption{We finetuned pretrained Qwen-2.5-VL-7B~\cite{qwen25} on a training set $D_2$ (for low-level artifacts). We applied Full Fine-Tuning (FFT) and LoRA-based adaptation to train ViT (V), Large Language Model (L), and ViT\&Large Language Model (V\&L). Our approach (with LoRA rank=64) achieves superior accuracy, demonstrating its capabilities in identifying generative artifacts.
  In contrast, standard LoRA tuning on the original MLLM architecture fails to capture these low-level artifacts effectively.}
  \label{fig:motivation}
\end{figure}

To investigate the deficiencies of MLLMs in AIGI detection, we selected the pretrained Qwen-2.5-VL-7B as the backbone and analyzed its characteristics on the datasets: $D_1$ (focusing on semantics) and $D_2$ (focusing on low-level generator traces).
Our analysis reveals that the pretrained MLLM, primarily optimizing vision-language alignment, naturally lacks sensitivity to the subtle low-level artifacts in $D_2$.
This creates a dilemma in representation learning: standard LoRA tuning on the original single-stream MLLM architecture is insufficient to recover these suppressed features, while full fine-tuning captures them but induces catastrophic forgetting of the original semantic knowledge (see Figure~\ref{fig:motivation} and Table~\ref{tab:commonbench}).
\begin{simplebox}
  \footnotesize\textbf{FINDING 1.} \textit{Native MLLMs cannot learn generalizable low-level artifacts without aggressively modifying the feature space, which compromises the extraction of semantic features.}
\end{simplebox}
Therefore, balancing the learning of low-level artifacts with the preservation of semantic features is key to efficient MLLM-based detection.
\begin{figure}
  \centering
  \includegraphics[width=\linewidth]{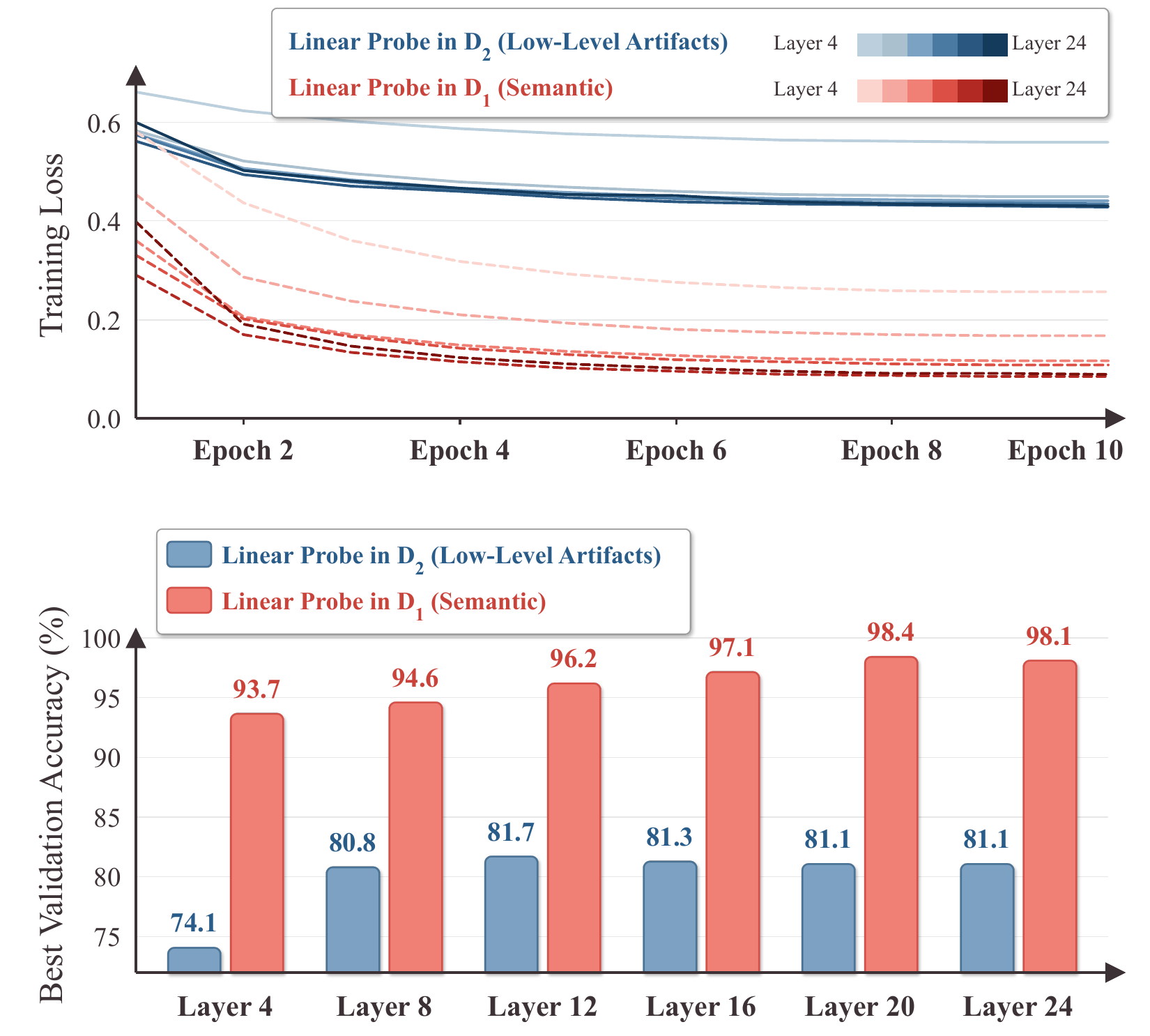}
  \caption{Layer-wise linear probing on Qwen-2.5-VL-7B. (Top) Training loss for $D_1$ (semantics-focused, dashed lines) converges faster and lower than for $D_2$ (artifact-focused, solid lines). (Bottom) Validation accuracy highlights a semantic convergence zone (layers 16–20) where $D_1$ performance peaks, while $D_2$ accuracy remains stagnant ($\simeq 81\%$) across all depths. }
  \label{fig:motivation3}
\end{figure}
To identify the specific layers responsible for semantic processing, we examined the layer-wise contribution to detection using linear probes on Qwen-2.5-VL-7B.
As shown in Figure~\ref{fig:motivation3}, detection accuracy for semantic anomalies ($D_1$) improves rapidly and converges in the early-to-middle layers (1-16), whereas artifact detection ($D_2$) remains ineffective throughout.
This indicates that these early-to-middle layers are the foundational "semantic convergence zone".
\begin{simplebox} 
  \footnotesize \textbf{FINDING 2.} \textit{The early-to-middle layers of Native MLLMs are critical for forming semantic representations.} 
\end{simplebox}
These findings explain why full fine-tuning compromises semantic capabilities: forcing the early layers to learn contradictory low-level artifact cues disrupts their established semantic convergence. Consequently, our strategy explicitly decouples this learning process. 
We preserve the early layers to maintain robust pre-formed semantic features and employ a Residual Injection strategy to introduce artifact features solely into the deep layers (post-convergence), where they can be integrated without interference.

\subsection{Model Architecture}
Our analysis (see Section~\ref{analysis}) indicates a fundamental limitation: pre-trained MLLMs lack the inherent sensitivity to low-level artifacts necessary for AIGI detection. 
Although full fine-tuning can enable this capability, it comes at a high cost, resulting in the catastrophic forgetting of the model's robust semantic representations.
To address the problem, we propose \textbf{Deep Visual Residual MLLM (Deep-VRM)}, a streamlined architecture designed to heighten sensitivity to low-level artifacts while preserving robust semantic reasoning. 
Given an input image $I$ and a text instruction sequence $\mathbf{x}_{1:N}$, the visual embeddings derived from the encoder $\mathcal{V}_o$ are concatenated with the textual embeddings to form the initial embedding sequence $\mathbf{H}^{(0)}$:
\begin{equation}
  \mathbf{H}^{(0)} = [\mathbf{v}_{\text{emb}}, \mathbf{x}_1, \dots, \mathbf{x}_N], \quad \text{where } \mathbf{v}_{\text{emb}} = \mathcal{V}_o(I).
  \label{eq:input}
\end{equation}
\noindent
Subsequently, this input is processed by the frozen shallow layers of the LLM (denoted as $\text{LLM}_{\text{pre}}$). We define $K$ as the first trainable LLM layer, i.e., the residual injection boundary. Thus, $\text{LLM}_{\text{pre}}$ contains layers $1$ to $K-1$ and yields the intermediate hidden states $\mathbf{H}^{(K-1)}$, which can be decomposed into visual and textual components:
\begin{equation}
\begin{aligned}
    \mathbf{H}^{(K-1)}
    &= \text{LLM}_{\text{pre}}(\mathbf{H}^{(0)}) \\
    &= \bigl[\mathbf{h}_v^{(K-1)}, \mathbf{h}_{t,1}^{(K-1)}, \dots, 
    &\quad \mathbf{h}_{t,N}^{(K-1)}\bigr].
\end{aligned}
\label{eq:pre_stage}
\end{equation}
Since these frozen early layers tend to suppress high-frequency artifact signals, we introduce a ``Green Road'' residual pathway for direct intervention. 
We equip the visual encoder $\mathcal{V}_o$ with lightweight LoRA adapters, denoting this adapted module as $\mathcal{V}_a$, to explicitly extract perceptual features from the image $I$ and inject them into the visual embeddings $\mathbf{h}_v^{(K-1)}$ before layer $K$. This residual connection effectively bypasses the semantic extraction stage of the early-to-middle layers:
\begin{equation}
    \mathbf{\tilde{h}}_v^{(K-1)} = \alpha \cdot \mathbf{h}_v^{(K-1)} + \beta \cdot \mathcal{V}_{a}(I),
    \label{eq:injection}
\end{equation}
where $\alpha$ and $\beta$ are scaling factors (set to $0.5$) that balance the semantic context with raw artifact cues. Finally, the enhanced visual tokens are re-concatenated with the original textual context and fed into the remaining layers ($\text{LLM}_{\text{post}}$) to generate the final response $\mathbf{O}$:
\begin{equation}
\begin{aligned}
    \mathbf{\tilde{H}}^{(K-1)}
    &= \bigl[\mathbf{\tilde{h}}_v^{(K-1)}, \mathbf{h}_{t,1}^{(K-1)}, \dots, 
    &\quad \mathbf{h}_{t,N}^{(K-1)}\bigr], \\
    \mathbf{O} &= \text{LLM}_{\text{post}}(\mathbf{\tilde{H}}^{(K-1)}).
\end{aligned}
\label{eq:post_stage}
\end{equation}

\begin{figure}
  \centering
  \includegraphics[width=1\linewidth]{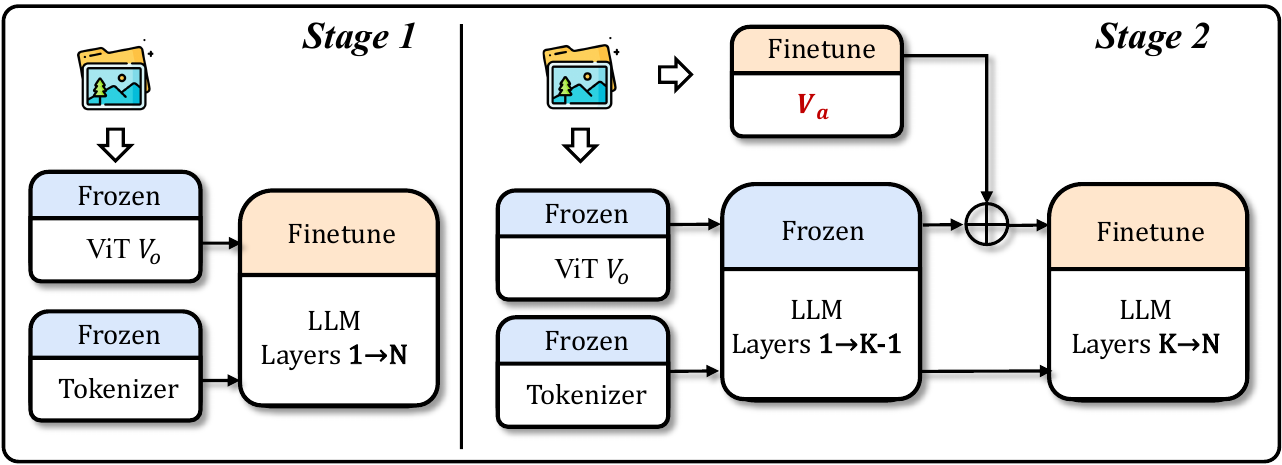}
  \caption{The architecture of our proposed method. The training process consists of two stages, where different parameters are finetuned in each stage. $\mathcal{V}_a$ is constructed by integrating an adapter into the base architecture of $\mathcal{V}_o$.}
  \label{fig:architecture}
\end{figure}

\subsection{Training Strategy}
We fine-tune the proposed MLLM using a standard auto-regressive loss. Given an input image $I$ and a target response sequence $Y = \{y_1, y_2, \dots, y_L\}$, the model is optimized by minimizing the negative log-likelihood:$$\mathcal{L}_{SFT} = -\sum_{i=1}^{L} \log P(y_i \mid I, y_{<i}; \Theta)$$where $\Theta$ denotes the set of trainable parameters. As illustrated in Figure ~\ref{fig:architecture}, our approach consists of two stages:

\paragraph{Stage 1:} Semantic Alignment and Prior Activation. In this stage, we employ $D_1$ to fine-tune the LLM component, while keeping all other modules frozen. This stage is designed to activate and align the LLM’s inherent pre-trained knowledge, enabling it to perform AIGI detection based on high-level semantic knowledge.
\paragraph{Stage 2:} Artifact-Aware Refinement. We integrate both $D_1$ and $D_2$ to train the residual injection architecture. To preserve the discriminative priors established in the first stage, the first $K-1$ LLM layers remain frozen. The residual features extracted by $\mathcal{V}_a$ are injected before layer $K$, after which $\mathcal{V}_a$ and the LLM layers from $K$ to $N$ are optimized. This design enables artifact-aware refinement while preserving the semantic representations formed by the frozen early layers.

\section{Experiments}

\paragraph{Implementation}
We employ Qwen-2.5-VL-7B~\cite{qwen25} as the default backbone. 
All images are resized to match an equivalent pixel budget of $512 \times 512$ while preserving their original aspect ratios during training. 
We train the model for 2 epochs using AdamW with $\beta_1=0.9$, $\beta_2=0.95$, and weight decay $1e^{-3}$, following a cosine learning-rate decay schedule. 
The learning rates are set to $1e^{-6}$ for the projector and $1e^{-4}$ for the visual encoder and LLM parameters. 
We apply LoRA~\citep{hu2022lora} with rank $r=64$ and LoRA alpha $128$. 
The adapted visual branch $\mathcal{V}_a$ is initialized from the frozen visual encoder $\mathcal{V}_o$ and equipped with LoRA adapters; therefore, compared with the standard baseline MLLM, our method introduces no additional trainable parameters beyond the LoRA parameters. 
For the default Qwen-2.5-VL-7B setting, the total number of trainable parameters is 115.02M.

\paragraph{Evaluation Metric and Comparison Methods}
In evaluation, following previous works~\cite{yan2025orthogonal,chen2025dual}, we use balanced accuracy for benchmarks containing real and fake images. For fake-only benchmarks, we report the fake-class accuracy.
To comprehensively verify the effectiveness of our method, we also provide the results of the following models:
Xception~\citep{chollet2017xception}, CNNSpot~\citep{wang2020cnn}, F3Net~\citep{qian2020thinking}, GramNet~\citep{liu2020global}, UnivFD~\citep{ojha2023towards}, NPR~\citep{tan2024rethinking}, AIDE~\citep{yan2025a}, DIRE~\citep{wang2023dire}, DRCT~\citep{chen2024drct}, OMAT~\citep{zhou2025breaking}, AIGI-Holmes~\citep{zhou2025aigi}, SAFE~\citep{li2025improving}, C2P-CLIP~\citep{tan2025c2p}, FatFormer~\citep{liu2024forgery}, CO-SPY~\citep{cheng2025co}, PatchShuffle~\citep{zheng2024breaking}, Forensic-Chat~\citep{lin2025seeing}.

\subsection{Evaluation for Generalization}
\paragraph{Performance on the AIGI detection benchmarks}
To verify the generalization of our method, we conduct experiments to evaluate the performance on the following datasets: GenImage~\citep{zhu2023genimage}, GenImage++~\cite{zhou2025breaking}, AIGI-Holmes~\cite{zhou2025aigi}, EvalGen~\cite{chen2025dual}, CommunityAI~\cite{park2025community}, SynthBuster~\cite{Synthbuster}, AIGI-NOW~\cite{chen2025task}. 
As detailed in Tables ~\ref{tab:genimage},~\ref{tab:synthbuster}, ~\ref{tab:compare-aiginow} and Table ~\ref{tab:genimage++},~\ref{tab:evalgen},~\ref{tab: comfor},~\ref{tab:aigi_holmes} in Appendix, our method consistently achieves SOTA performance. This consistent superiority across diverse generative sources underscores the effectiveness of Deep-VRM. 

\begin{table}[t]
\centering
\caption{The generalizable performance (ACC (\%)) in GenImage Dataset~\citep{zhu2023genimage}.} 
\resizebox{\linewidth}{!}{
\begin{tabular}{l|cccccccc|c}
\toprule
Model & MidJ & SDv1.4 & SDv1.5 & ADM   & GLIDE & Wukong & VQDM  & BigGAN & \textbf{AVG} \\
\midrule
Xception & 57.97 & 98.06 & 97.98 & 51.16 & 57.51 & 97.79 & 50.34 & 48.74 & 69.94 \\
CNNSpot & 61.25 & 98.13 & 97.54 & 51.50 & 55.13 & 93.51 & 51.83 & 51.06 & 69.99 \\
F3Net & 52.26 & 99.30 & 99.21 & 49.64 & 50.46 & 98.70 & 45.56 & 49.59 & 68.09 \\
GramNet & 63.00 & 94.19 & 94.22 & 48.69 & 46.19 & 93.79 & 49.20 & 44.71 & 66.75 \\
UnivFD & 77.29 & 97.01 & 96.67 & 50.94 & 78.47 & 91.52 & 65.72 & 55.91 & 76.69 \\ 
NPR & 62.00 & \textbf{99.75} & 99.64 & 56.79 & 82.69 & 97.89 & 54.43 & 52.26 & 75.68 \\
AIDE & 79.38 & 99.74 & \textbf{99.76} & 78.54 & 91.82 & 98.65 & 80.26 & 66.89 & 86.88 \\
DIRE  & 51.11 & 55.07 & 55.31 & 49.93 & 50.02 & 53.71 & 49.87 & 49.85 & 51.86 \\
DRCT/Conv-B &{94.43} & 99.37 & 99.19 & 66.42 & 73.31 &{99.25} & 76.85 & 59.41 & 83.53 \\
DRCT/UniFD   & 85.82 & 92.33 & 91.87 & 75.18 & 87.44 & 92.23 & 89.12 & 87.38 & 87.67 \\
OMAT &90.36 &97.52 &97.46 &83.82 &97.41 &97.62 &{95.53} &{97.34} &94.63 \\
\midrule
\rowcolor{blue!3} \textbf{Deep-VRM} &\textbf{96.16} &99.65 &99.42 &\textbf{89.82} &\textbf{99.08} &\textbf{99.44} &\textbf{97.29} &\textbf{98.50} &\textbf{97.42} \\
\bottomrule
\end{tabular}
}
\label{tab:genimage}
\end{table}

\begin{table}
  \centering
  \caption{Generalizable performance (ACC (\%)) on evaluation set in SynthBuster~\citep{Synthbuster}. The methods marked with $^{*}$ indicate results reported in the original paper. This dataset includes fake images only.}
  \label{tab:synthbuster}
\resizebox{\linewidth}{!}{
  \begin{tabular}{l|cccccccccccc}
    \toprule
    Method & Glide &SD1.3 &SD1.4 &SD2 &SD XL &MidJ &DALLE2 &DALLE3 &Firefly &\textbf{AVG} \\
    \midrule
    UnivFD* &10.10 &24.30 &21.80 &34.40 &21.50 &0.00 &42.40 &0.00 &61.70 &14.30 \\
    C2P-CLIP &12.00 &51.10 &54.20 &39.10 &56.20 &6.50 &12.00 &27.20 &19.70 &30.89 \\
    DeeCLIP &48.30 &93.30 &93.40 &68.16 &54.10 &30.00 &89.44 &0.40 &{71.50} &60.96 \\
    DRCT &14.10 &89.60 &88.20 &\textbf{99.90} &{89.60} &\textbf{99.40} &4.10 &35.60 &11.40 &59.10 \\
    PatchShuffle &80.40 &100.00 &100.00 &82.49 &77.70 &82.80 &19.80 &8.10 &13.50 &62.75 \\
    \midrule
    \rowcolor{blue!3}  \textbf{Deep-VRM} &\textbf{99.90} &\textbf{99.80} &\textbf{100.00} &{71.90} &\textbf{95.80} &{96.80} &\textbf{99.90} &\textbf{99.90} &\textbf{86.50} &\textbf{94.50} \\
    \bottomrule
  \end{tabular}
}
\end{table}

\begin{table}[t]
\centering
\caption{Performance (ACC (\%)) on WildRF~\citep{cavia2024real} and AIGI-Bench~\citep{li2025aigi}, two benchmarks designed to reflect wild scenario and evaluate the robustness of detectors. }
\label{tab:wild}
\resizebox{\linewidth}{!}{
\begin{tabular}{l|cccccccccc}
\toprule
\multirow{2}{*}{Method} & \multicolumn{4}{c}{\textit{WildRF}} & \multicolumn{3}{c}{\textit{AIGI Bench}}\\
\cmidrule(lr){2-5}\cmidrule(lr){6-8}
 & FaceBook & Reddit & Twitter & \textbf{AVG} & SocialRF & CommunityAI & \textbf{AVG} \\
\midrule
FatFormer & 64.38  & 76.65 & 40.00 & 60.34 &57.98 &50.62 &54.30 \\
CO-SPY    & 50.00  & 56.79 & 73.30 & 60.03 &55.54 &53.02 &54.28 \\
C2P-CLIP  & 54.38  & 68.40 & 47.27 & 56.68  &53.13 &50.98 &52.06 \\
SAFE      & 62.50  & 61.70 & 40.33 & 54.84 &58.00 &54.25 &56.13 \\
AIDE      & 75.00  & 55.48 & 48.00 & 59.49 &57.80 &54.15 &55.98 \\
\midrule
\textbf{Deep-VRM} & \textbf{85.94} & \textbf{91.07} & \textbf{91.53} & \textbf{89.51} &\textbf{83.68} &\textbf{97.28} &\textbf{90.48} \\ 
\bottomrule
\end{tabular}
}
\end{table}

\begin{table}[t]
  \centering
  
  \caption{Ablation Study to explore the impact to vary the insert depth $K$ of visual residual features.}
  \label{tab:ablation_depth}
  \resizebox{\linewidth}{!}{
  \begin{tabular}{l|cccc|c}
    \toprule
    \textbf{Depth} &GenImage &SynthBuster &AIGI-Bench &WildRF &\textbf{AVG} \\
    \midrule
    4 &97.05 &\textbf{96.69} &83.68 &74.82 &88.06 \\
    8 &96.83 &94.99 &86.65 &74.68 &88.29 \\
    12 &97.26 &93.40 &\textbf{91.06} &89.12 &92.91 \\
    16 &\textbf{97.39} &94.50 &90.25 &\textbf{89.51} &\textbf{92.97} \\
    20 &96.42 &94.44 &90.30 &89.20 &92.59 \\
    \bottomrule
  \end{tabular}
  }
\end{table}

\begin{table*}[tb!]
\centering
\caption{The generalizable performance (ACC (\%)) in AIGI-Now Dataset~\cite{chen2025task}.}
\label{tab:compare-aiginow}

\newcommand{\sem}{\textit{Sem}}
\newcommand{\pix}{\textit{Pix}}
\resizebox{\textwidth}{!}{
\begin{tabular}{l|cccccccccccccccccc|cccc}
\toprule
& \multicolumn{2}{c}{Nano Banana}
& \multicolumn{2}{c}{GPT-4o} 
& \multicolumn{2}{c}{Jimeng} 
& \multicolumn{2}{c}{Kling} 
& \multicolumn{2}{c}{Minimax} 
& \multicolumn{2}{c}{Flux Pro}
& \multicolumn{2}{c}{Flux Krea}
& \multicolumn{2}{c}{Flux Dev}
& \multicolumn{2}{c}{Flux Kontext}
& \multicolumn{2}{c}{\textbf{AVG}} \\
\cmidrule(lr){2-3}\cmidrule(lr){4-5}\cmidrule(lr){6-7}\cmidrule(lr){8-9}\cmidrule(lr){10-11}
\cmidrule(lr){12-13}\cmidrule(lr){14-15}\cmidrule(lr){16-17}\cmidrule(lr){18-19}\cmidrule(lr){20-21}
Method 
& \pix & \sem & \pix & \sem & \pix & \sem & \pix & \sem 
& \pix & \sem & \pix & \sem & \pix & \sem & \pix & \sem 
& \pix & \sem & \pix & \sem \\
\midrule
NPR
& 84.4 & 49.9 & 91.2 & 50.1 & 41.9 & 49.9 & 88.3 & 49.9 & 52.1 & 50.0
& 42.7 & 50.0 & 52.3 & 50.0 & 89.3 & 50.0 & 79.2 & 49.9
& 69.0 & 50.0 \\
UnivFD
& 51.0 & 51.4 & 53.8 & 53.3 & 50.7 & 52.3 & 51.4 & 53.9 & 50.5 & 52.1 
& 51.9 & 52.0 & 51.0 & 52.2 & 63.6 & 49.9 & 57.5 & 55.9 
& 52.0 & 52.6 \\
FatFormer
& 53.4 & 49.9 & 49.9 & 49.8 & 48.4 & 49.9 & 54.8 & 50.0 & 49.1 & 50.0 
& 48.4 & 50.0 & 49.5 & 50.0 & 52.4 & 50.0 & 63.6 & 49.9 
& 52.1 & 49.9 \\
SAFE
& {99.3} & 50.1 & {99.7} & 50.4 & 50.0 & 50.5 & {99.6} & 50.0 & 49.8 & 50.1 
& 49.9 & 50.5 & 51.2 & 50.1 & \textbf{99.8} & {75.3} & \textbf{98.8} & \textbf{83.4} 
& 77.5 & 56.7\\
C2P-CLIP
& 50.0 & 50.0 & 49.8 & 50.0 & 49.8 & 50.0 & 50.7 & 49.9 & 50.3 & 50.0 
& 52.0 & 50.0 & 50.1 & 50.0 & 52.6 & 50.0 & 60.7 & 50.0 
& 51.7 & 50.0 \\
AIDE
& {98.9} & 51.8 & 74.7 & 53.5 & 63.9 & 51.4 & {98.2} & 55.4 & 51.4 & 54.1 
& 60.1 & 53.8 & 50.4 & 56.9 & {99.1} & 59.0 & {97.9} & {80.6} 
& 77.1 & 57.4 \\
DRCT
& 68.9 & 57.7 & 75.0 & 57.0 & {89.3} & 57.7 & 66.9 & 56.8 & 78.4 & 57.1 
& {83.0} & 58.2 & {91.6} & 58.2 & 86.5 & 58.6 & 86.9 & 55.5 
& {80.7} & 57.4 \\
AlignedForensics
& 88.0 & 50.3 & 50.8 & 50.8 & 82.1 & 50.0 & 69.6 & 50.2 & 56.7 & 50.3 
& 65.5 & 49.9 & 59.5 & 50.1 & 78.6 & 50.2 & 65.0 & 50.0 
& 68.4 & 50.2 \\
CO-SPY
& 74.1 & {58.3} & 78.9 & {63.1} & 83.0 & {76.2} & 88.3 & {78.8} & {77.9} & {65.6} 
& 77.3 & {72.7} & {89.1} & {74.0} & 88.5 & 74.9 & 65.1 & 66.0 
& 80.2 & {70.0} \\
BFree
& 69.6 & 52.8 & 59.2 & 56.7 & 75.6 & 54.5 & 86.1 & 60.7 & 56.2 & 49.9 
& 71.8 & 54.5 & 59.5 & 51.5 & 74.8 & 54.5 & 77.1 & 53.4 
& 70.0 & 54.3 \\
\midrule
\rowcolor{blue!3} \textbf{Deep-VRM} &\textbf{99.9} &\textbf{99.0} &\textbf{100.0} &\textbf{97.0} &\textbf{99.4} &\textbf{96.5} &\textbf{99.9} &\textbf{97.3} &\textbf{93.6} &\textbf{92.4} &\textbf{96.2} &\textbf{97.3} &\textbf{94.4} &\textbf{88.5} &{99.1} &\textbf{96.3} &{96.7} &79.60 &\textbf{97.7} &\textbf{93.7} \\
\bottomrule
\end{tabular}
}
\end{table*}

\subsection{Evaluation for Performance in Wild Scenario}
Considering practical applications, validation on 'in-the-wild' datasets is necessary. Real-world images often suffer from complex post-processing (e.g., re-compression, resizing) and originate from diverse, unknown sources. To address this, we selected WildRF~\citep{cavia2024real} and AIGI-Bench~\citep{li2025aigi}, which comprise data collected from social media platforms (e.g., Facebook, Reddit, Twitter) and online communities. As shown in Table~\ref{tab:wild}, our method consistently outperforms SOTA competitors by a significant margin. 
This validates our model's robustness and its ability to generalize effectively to unconstrained scenarios where low-level artifacts are often degraded.

\subsection{Comparison for other MLLMs}
To verify the effectiveness of our proposed method, we conducted a comprehensive comparison against representative MLLMs. 
The baselines include general-purpose models (e.g., Qwen-2-VL~\cite{wang2024qwen2}, Qwen-2.5-VL~\cite{qwen25}, LLaVA-1.5~\cite{liu2023visual}) and a specialized forensic MLLM (Forensic-Chat) tailored for AI-generated image detection.
In Table~\ref{tab:mllm_comparison}, it can be observed that general-purpose MLLMs struggle with the detection task, yielding an average accuracy ranging from 50\% to 68\%. 
While the domain-specific Forensic-Chat~\cite{lin2025seeing} achieves a respectable average accuracy of 88.75\%, our method achieves the best performance with an average accuracy of \textbf{93.28\%}. 
Notably, on the challenging benchmarks AIGI-Bench and WildRF, our method surpasses the strongest baseline by margins of 8.29\% and 5.57\%, respectively. 
This demonstrates that injecting visual residual features effectively bridges the gap between general understanding and low-level artifact detection.

\begin{table}[h]
  \scriptsize
  \centering
  \caption{Comparison of performance (ACC\%) across different MLLMs on AIGI detection benchmarks. Forensic-Chat is an MLLM-based detector tailored for AIGI detection.}
  \resizebox{\linewidth}{!}{
  \begin{tabular}{l|cccc|c}
    \toprule
    \textbf{Method} &\textbf{GenImage} &\textbf{AIGI-Now} &\textbf{AIGI Bench} &\textbf{WildRF} &\textbf{AVG} \\
    \midrule
    Qwen-2-VL-2B &56.54 &56.66 &57.97 &62.42 &58.40 \\
    Qwen-2-VL-7B  &56.66 &67.56 &66.43 &83.94 &68.65 \\
    Qwen-2.5-VL-3B &63.93 &62.38 &69.48 &78.69 &68.62 \\
    Qwen-2.5-VL-7B &50.18 &57.54 &53.78 &54.64 &54.04 \\
    LLaVA-1.5-7B &49.84 &50.17 &50.89 &51.41 &50.58 \\
    \midrule
    Forensic-Chat &97.55 &94.21 &82.19 &81.08 &88.75 \\
    \rowcolor{blue!3} \textbf{Deep-VRM} &97.42 &95.70 &90.48 &89.51 &93.28 \\
    \bottomrule
  \end{tabular}
  }
  \label{tab:mllm_comparison}
\end{table}

\subsection{Ablation Study}

\paragraph{Impact of Visual Residual Depth.}
We analyze the impact of feature injection depth on model performance (Table~\ref{tab:ablation_depth}).
The results corroborate the representation conflict hypothesis discussed in Sec.~\ref{sec:intro}.
Shallow injection (Depth 4) forces the model to prioritize low-level artifacts prematurely, which boosts accuracy on specific uncompressed datasets (e.g., SynthBuster) but disrupts the early encoding of semantic features, leading to poor generalization on compressed data (e.g., WildRF).
Conversely, deep injection (Depth 16) preserves the integrity of the semantic latent space in the early layers while introducing forensic signals at a stage where the model can perform joint reasoning.
This configuration yields a significant robustness gain (+14\% on WildRF compared to Depth 4) while maintaining competitive precision on pixel-level artifacts, motivating our choice of Depth 16 as the default setting.

\paragraph{Ablation on Different Training Stages}
We conducted experiments to investigate the individual contributions of different training stages to the overall performance. Table~\ref{tab:ablation_stage} compares the results of the initial training phase (Stage 1) against the full training pipeline (Stage 1 + Stage 2). Compared with the first stage alone, the integration of Stage 2 yields a significant performance gain.

\begin{table}[t]
  \centering
  \tiny
  \caption{Ablation Study to explore the impact of different stages.}
  \label{tab:ablation_stage}
  \begin{tabular}{l|cccc|c}
    \toprule
    \textbf{Model} &GenImage &WildRF &AIGI-Bench &SynthBuster &\textbf{AVG} \\
    \midrule
    Stage 1 &95.34 &73.27 &69.55 &79.86 &79.51 \\
    Stage 1 + Stage 2 &97.42 &89.51 &90.48 &94.50 &92.98 \\
    \bottomrule
  \end{tabular}
\end{table}

\begin{table}[t]
  \centering
  
  \caption{Ablation Study on Freezing the Early LLM Layers.}
  \label{tab:ablation_frozen}
  \resizebox{\linewidth}{!}{
  \begin{tabular}{l|cccc|c}
    \toprule
    \textbf{Freezing} &GenImage &WildRF &AIGI-Bench &SynthBuster &\textbf{AVG} \\
    \midrule
    No &97.95 &83.49 &87.39 &98.62 &91.86 \\
    Yes (Ours) &97.42 &89.51 &90.48 &94.50 &92.98 \\
    \bottomrule
  \end{tabular}
  }
\end{table}

\paragraph{Ablation on Freezing Early LLM Layers}
For our new architecture, we investigate the effect of freezing early LLM layers versus fine-tuning all layers. The results in Table~\ref{tab:ablation_frozen} demonstrate that while fine-tuning all layers (No) performs comparably on standard datasets, our strategy of freezing early layers (Yes) significantly boosts performance on challenging in-the-wild benchmarks such as WildRF and AIGI-Bench. Consequently, our method achieves a higher average accuracy of 92.98\%, validating that freezing early layers effectively prevents the forgetting of general semantic knowledge required for robust detection.

\begin{table}[t]
  \centering
  \caption{Ablation Study on altering different backbones in our method.}
  \label{tab:backbone}
  \resizebox{\linewidth}{!}{
  \begin{tabular}{l|cccc|c}
    \toprule
    \textbf{Model} &GenImage &WildRF &AIGI-Bench &SynthBuster &\textbf{AVG} \\
    \midrule
    Qwen-2-VL-7B &{97.28} &85.41 &89.11 &96.08 &91.97 \\
    LLaVA-1.5-7B &92.02 &\textbf{92.47} &\textbf{92.48} &\textbf{96.48} &\textbf{93.36} \\
    Qwen-2.5VL-3B &85.72 &71.52 &80.80 &92.54 &82.65 \\
    Qwen-2.5VL-7B  &\textbf{97.42} &89.51 &90.48 &94.50 &92.98 \\
    \bottomrule
  \end{tabular}
  }
\end{table}
\paragraph{Ablation on Altering Backbone}
To evaluate the generalization capability of our proposed method, we apply it across various MLLM architectures, including Qwen-2.5-VL-7B, Qwen-2-VL-7B~\cite{wang2024qwen2}, and LLaVA-1.5-7B~\cite{liu2023visual}. As shown in Table~\ref{tab:backbone}, our approach consistently delivers robust performance across multiple benchmarks, demonstrating its architecture-agnostic effectiveness. However, we observe a performance degradation in the Qwen-2.5-VL-3B model. We attribute this primarily to its reduced parameter scale, which limits its capacity to capture complex semantic features. Furthermore, our use of a fixed LoRA rank ($r=64$) may have induced underfitting, as the trainable parameter budget decreases proportionally with the model's hidden dimension.

\paragraph{Ablation on Residual Fusion Coefficients}

{To analyze the impact of the residual fusion coefficients, we vary $\alpha$ and $\beta$ in Eq.~\ref{eq:injection}. In our implementation, $\alpha$ scales the pre-formed intermediate visual-token representation $\mathbf{h}_v^{(K-1)}$, while $\beta$ scales the injected artifact-specific features $\mathcal{V}_a(I)$. As shown in Table~\ref{tab:ablation_coeff}, the balanced setting $(\alpha,\beta)=(0.5,0.5)$ achieves the best average performance. When $\alpha$ is very small, the pre-formed representation is strongly down-weighted, leading to clear drops, especially on GenImage and SynthBuster. This suggests that preserving a sufficient intermediate representation is important for stable fusion. However, once $\alpha$ reaches $0.5$, further increasing its value brings limited additional gains, and the average performance becomes nearly saturated. This indicates that the model is relatively robust to moderate changes in the fusion coefficients, likely because the trainable modules can partially adapt to different feature scales during optimization. These observations suggest that the fusion coefficients mainly affect the relative scale of the two branches, rather than imposing a structural constraint as the injection depth does. Thus, this ablation should be viewed as a sensitivity analysis of the fusion design, showing that $(\alpha,\beta)=(0.5,0.5)$ offers a simple and effective trade-off between preserving the pre-formed semantic representation and introducing artifact-aware residual cues.}

\begin{table}[t]
  \centering
  \caption{Ablation study on the residual fusion coefficients $\alpha$ and $\beta$ in Eq.~\ref{eq:injection}.}
  \label{tab:ablation_coeff}
  \resizebox{\linewidth}{!}{
  \begin{tabular}{c|cccc|c}
    \toprule
    $(\alpha, \beta)$ & GenImage & SynthBuster & AIGI-Bench & WildRF & \textbf{AVG} \\
    \midrule
    $(0.1, 0.9)$ & 94.94 & 84.42 & 90.67 & 89.04 & 89.77 \\
    $(0.3, 0.7)$ & 94.29 & 90.02 & \textbf{91.48} & \textbf{90.31} & 91.53 \\
    $(0.5, 0.5)$ & \textbf{97.39} & \textbf{94.50} & 90.25 & 89.51 & \textbf{92.91} \\
    $(0.7, 0.3)$ & 96.76 & 94.24 & 90.10 & 86.97 & 92.76 \\
    $(0.9, 0.1)$ & 96.78 & 92.82 & 91.21 & 90.04 & 92.71 \\
    \bottomrule
  \end{tabular}
  }
\end{table}

\paragraph{Impact for Dataset}
To decouple the influence of training data, we retrained representative expert detection methods (UnivFD and AIDE) and the baseline MLLM (Qwen-2.5-VL-7B) using the combined datasets $D_1$ and $D_2$. For the baseline MLLM, we employ LoRA fine-tuning with a rank of 64.
These models were subsequently evaluated on diverse real-world benchmarks to assess their generalization capabilities. As presented in Table \ref{tab:ablation_data}, even when provided with the same high-quality forensic data, Deep-VRM significantly outperforms existing approaches across all metrics, particularly on ``in-the-wild" datasets like WildRF. This confirms that the performance gains are primarily attributed to our deep residual injection architecture rather than the dataset scale alone. 
\begin{table}[t]
  \centering
  \caption{Performance comparison of expert detection methods, the baseline MLLM, and ours when trained on the combined $D_1$ and $D_2$ datasets.}
  \label{tab:ablation_data}
  \resizebox{\linewidth}{!}{
  \begin{tabular}{l|ccccclccccccc}
    \toprule
    Method &GenImage &WildRF &AIGI-Bench &SynthBuster &\textbf{AVG} \\
    \midrule
    UnivFD &78.10 &70.15 &82.15 &53.15 &70.89 \\
    AIDE &93.19 &78.93 &82.14 &87.02 &85.32 \\
    \midrule
    Qwen-2.5-VL-7B &92.07 &70.85 &55.22 &69.99 &72.03 \\
    \midrule
    \rowcolor{blue!3}  \textbf{Deep-VRM} &\textbf{97.42} &\textbf{89.51} &\textbf{90.48} &\textbf{94.50} &\textbf{92.98} \\
    \bottomrule
  \end{tabular}
  }
\end{table}


\begin{table}[t]
  \centering
\caption{Validation of different feature fusion paradigms. Feature-Concat uses a linear classifier, Input Fusion follows the external expert paradigm by directly inserting auxiliary visual tokens into the input sequence, and Early/Late Fusion correspond to residual injection at shallow/deep layers.}
  \label{tab:fusion_paradigm}
  \resizebox{\linewidth}{!}{
  \begin{tabular}{l|cccc|c}
    \toprule
    \textbf{Method} & \textbf{GenImage} & \textbf{SynthBuster} & \textbf{AIGI-Bench} & \textbf{WildRF} & \textbf{AVG} \\
    \midrule
    Feature-Concat & 50.59 & 5.38 & 51.58 & 48.71 & 39.07 \\
    Input Fusion & 96.93 & 95.71 & 84.81 & 79.88 & 89.33 \\
    Early Fusion ($K{=}4$) & 97.05 & \textbf{96.69} & 83.68 & 74.82 & 88.06 \\
    Late Fusion ($K{=}20$) & 96.42 & 94.44 & \textbf{90.30} & 89.20 & 92.59 \\
    \rowcolor{blue!3} \textbf{Deep-VRM ($K{=}16$)} & \textbf{97.39} & 94.50 & 90.25 & \textbf{89.51} & \textbf{92.91} \\
    \bottomrule
  \end{tabular}
  }
\end{table}

\begin{figure*}
  \centering
  \includegraphics[width=\textwidth]{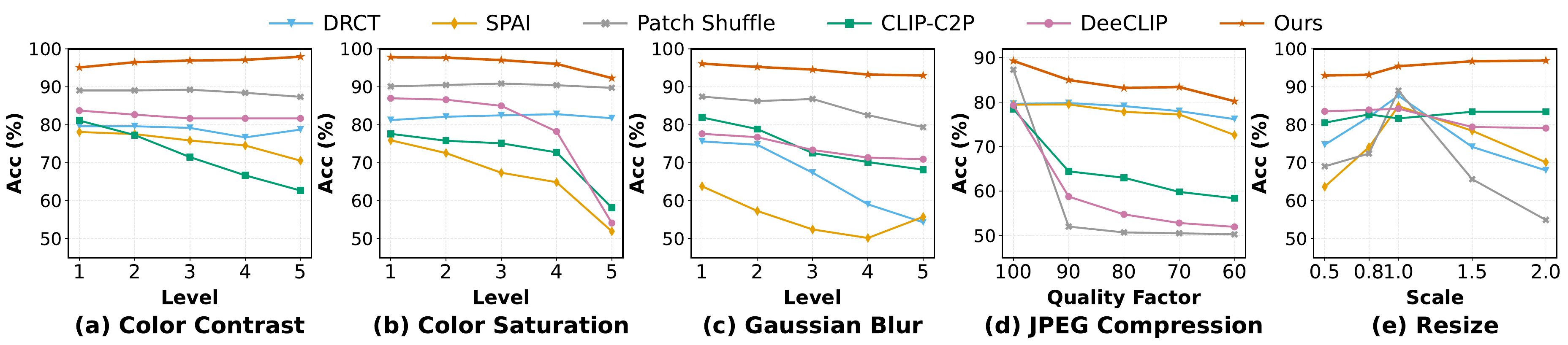}
  \caption{Robust performance (Acc (\%)) on GenImage after post-process. We deploy 5 different post-process to verify the robustness of our method and compared methods.}
  \label{fig:robust}
\end{figure*}

\paragraph{Comparison with Different Feature Fusion Paradigms.}
To isolate the source of Deep-VRM's improvement, we compare it with several variants that use the same adapted visual branch $\mathcal{V}_{a}$ but integrate it into the MLLM in different ways, as shown in Table~\ref{tab:fusion_paradigm}.
This comparison is intended to examine whether our gains come from the proposed deep residual injection design, rather than from merely introducing extra visual features.
First, we replace our residual injection with a Feature-Concat baseline, where the features from $\mathcal{V}_{a}$ are pooled and concatenated with the final LLM hidden state for classification:
\begin{equation}
    \hat{y}=\mathrm{MLP}\bigl([\mathbf{h}_{L};\operatorname{Pool}(\mathcal{V}_{a}(I))]\bigr).
    \label{eq:feature_concat}
\end{equation}
Although this baseline has access to the same adapted visual branch, its performance drops substantially, indicating that simply attaching artifact-aware features to a final classifier cannot effectively couple low-level forensic cues with the MLLM representation.
We then compare Deep-VRM with Input Fusion, which follows the common external expert paradigm by treating $\mathcal{V}_{a}$ as an auxiliary visual expert and directly inserting its unpooled visual tokens into the input sequence:
\begin{equation}
    \mathbf{H}^{(0)}_{\mathrm{IF}}
    = [\mathcal{V}_{o}(I), \mathcal{V}_{a}(I), \mathbf{x}_1, \dots, \mathbf{x}_N],
    \label{eq:input_fusion}
\end{equation}
where $\mathcal{V}_{a}(I)$ has the same token format as $\mathcal{V}_{o}(I)$ and is inserted without pooling.
Compared with Deep-VRM, this design exposes all subsequent LLM layers to additional expert tokens and therefore substantially increases the input sequence length, computation, and memory cost.
However, despite this heavier token-level fusion, Input Fusion remains clearly inferior to Deep-VRM, especially on the in-the-wild benchmarks WildRF and AIGI-Bench.
This shows that the advantage of Deep-VRM is not simply due to providing more expert-like visual tokens to the MLLM; rather, the location and manner of feature integration are critical.
Finally, we evaluate whether the residual pathway should be injected earlier or later than our default setting.
Early Fusion ($K{=}4$) injects artifact features before stable semantic representations are formed and performs poorly on WildRF, suggesting that premature intervention can disturb semantic processing.
Late Fusion ($K{=}20$) avoids such early interference but leaves fewer subsequent layers for joint reasoning over semantic and artifact-aware signals.
By injecting $\mathcal{V}_{a}(I)$ at $K{=}16$, Deep-VRM achieves the best average accuracy, supporting our design choice of introducing artifact-aware residual features after semantic convergence while still preserving sufficient depth for multimodal reasoning.


\subsection{Evaluation for Robustness}
In this experiment, we evaluate the robustness of Deep-VRM on GenImage under common image degradations, with results shown in Figure~\ref{fig:robust}. 
Since real-world images are often post-processed during uploading, sharing, and storage, such perturbations can weaken fragile pixel-level traces and challenge practical AIGI detection. 
We consider five representative degradations, including Color Contrast, Color Saturation, Gaussian Blur, JPEG compression, and Resize, covering both photometric changes and spatial/frequency distortions. 
Compared with existing models, Deep-VRM maintains consistently higher accuracy across degradation types and severity levels. 
This suggests that Deep-VRM does not rely on a single fragile artifact pattern, but jointly exploits low-level cues and semantic detection features through artifact-aware residual injection. 
Detailed degradation configurations are provided in the Appendix.

\subsection{Visualization}
\begin{figure}
  \centering
  \includegraphics[width=\linewidth]{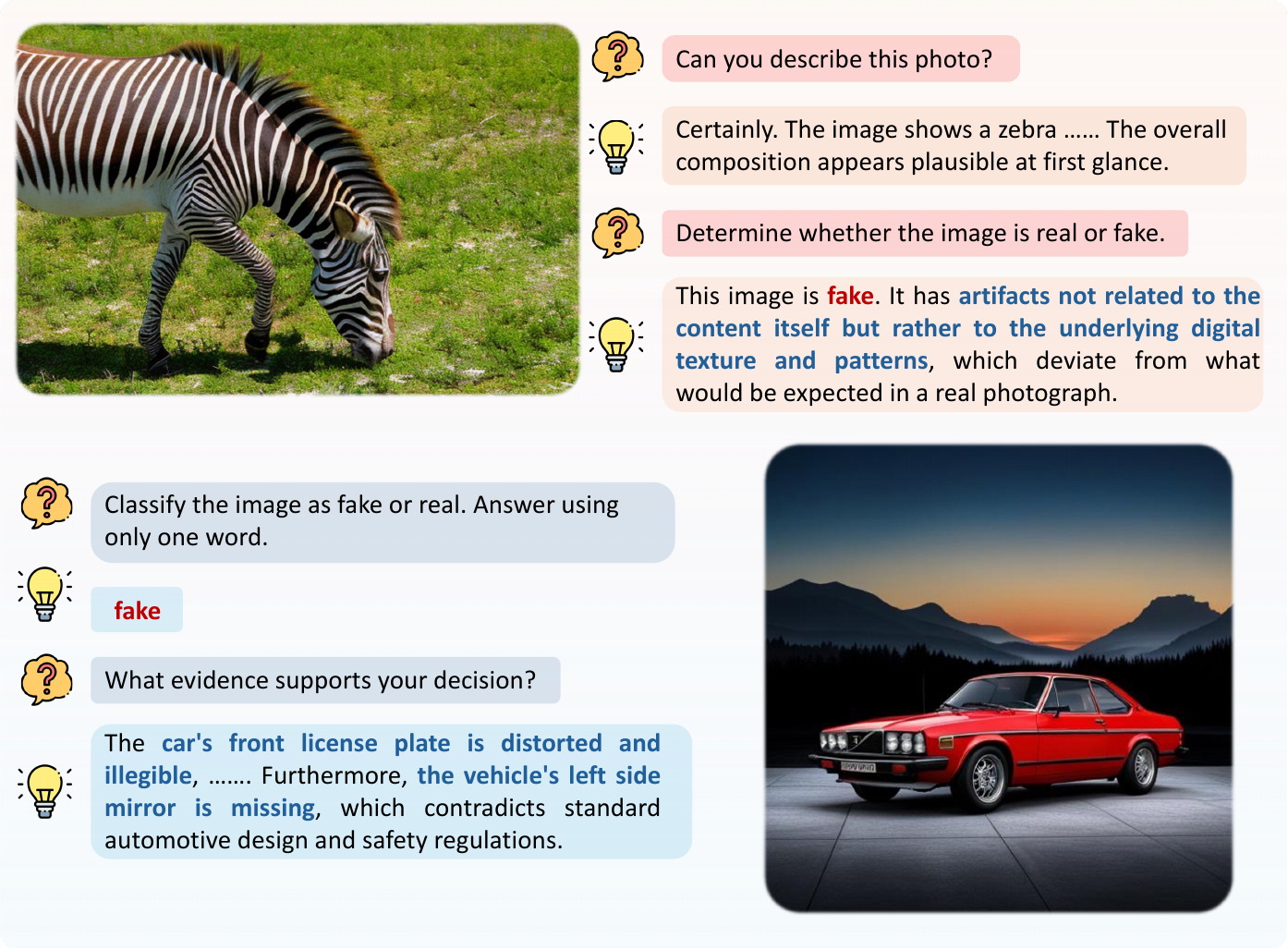}
  \caption{Visualization of conversational AIGI detection. Deep-VRM pinpoint low-level artifacts (e.g., digital textures in the zebra image) and semantic-level flaws (e.g., the missing mirror and distorted plate in the car image).}
  \label{fig:chat}
\end{figure}

Although Deep-VRM is not designed to enhance explainability, we visualize conversational examples of Deep-VRM in Figure~\ref{fig:chat}. The results demonstrate that our model does not merely provide binary classification but effectively articulates multi-level generative traces. In these open-ended responses, the model grounds its judgment in observable visual evidence rather than relying only on the final ``real'' or ``fake'' token.
Specifically, as shown in the zebra example, the model identifies low-level artifacts rooted in abnormal digital textures. Conversely, in the vehicle example, it detects semantic-level anomalies that violate real-world logic, such as the distorted license plate and the missing side mirror. This confirms that our strategy enables the MLLM to autonomously prioritize crucial features—whether subtle textures or complex semantic inconsistencies—to provide evidence-based detection.

\section{Conclusion}
In this paper, we presented Deep Visual Residual MLLM (Deep-VRM), a novel architecture designed to achieve full-spectrum forensic signal perception by resolving the representation conflict between semantic reasoning and low-level artifact detection in MLLMs. Our layer-wise analysis reveals a critical functional stratification: while early-to-middle layers of pretrained MLLMs are essential for semantic convergence, forcing them to learn generator-specific traces leads to catastrophic forgetting of semantic knowledge.
To address this, Deep-VRM employs a Deep Residual Injection strategy that preserves the frozen semantic processing of early layers while injecting artifact-specific visual features directly into the intermediate layers of the LLM. Extensive evaluations across diverse benchmarks—including GenImage, AIGI-Now, and challenging "in-the-wild" datasets like WildRF—demonstrate that our method achieves state-of-the-art performance without relying on external expert detectors. Furthermore, Deep-VRM exhibits superior robustness to real-world image degradations such as compression and resizing. By effectively bridging the gap between high-level semantic analysis and low-level forensic perception, Deep-VRM provides a unified and generalizable framework for reliable AI-generated image detection.

\newpage
\clearpage
\section*{Acknowledgement}
Dr Bin Li was supported in part by NSFC and in part by Shenzhen R\&D Program (Grant JCYJ20250604181211016, SYSPG20241211174032004).

\section*{Impact Statement}
This paper presents work whose goal is to advance the field of Machine
Learning. There are many potential societal consequences of our work, none
which we feel must be specifically highlighted here.
\bibliography{reference}
\bibliographystyle{icml2026}

\clearpage

\appendix

\section{Experiments}

\paragraph{Evaluation Setting of Deep-VRM}
In this paper, we conducted all performance evaluations using binary-choice prompts. Specifically, we standardized the input textual prompt as: \texttt{Is this image real or fake? Please just answer "real" or "fake"}. The model's accuracy was then calculated by parsing the generated output for the target tokens ``real'' or ``fake''.

\paragraph{Comparison with Other PEFT Methods}
To further examine whether the improvement comes merely from the choice of parameter-efficient fine-tuning (PEFT) method, we compare the standard LoRA baseline with OLoRA under the same rank setting ($r=64$). All variants are trained on $D_2$ and evaluated on GenImage. As shown in Table~\ref{tab:peft_appendix}, OLoRA improves over the standard LoRA baseline, but it still largely trails Deep-VRM. This indicates that mitigating parameter-space interference at the PEFT level alone is insufficient for learning low-level forensic artifacts in MLLMs. In contrast, Deep-VRM explicitly changes the feature integration pathway by injecting artifact-aware residual visual features after semantic convergence, leading to substantially stronger generalization.

\begin{table}[h]
  \centering
  \caption{Ablation study of different PEFT methods. All methods are trained on $D_2$ with rank $r=64$ and evaluated on GenImage.}
  \label{tab:peft_appendix}
  \begin{tabular}{l|c}
    \toprule
    \textbf{Method} & \textbf{GenImage} \\
    \midrule
    LoRA & 51.85 \\
    OLoRA & 60.40 \\
    Ours (LoRA) & \textbf{96.25} \\
    \bottomrule
  \end{tabular}
\end{table}

\paragraph{Performance on More AIGI detection benchmarks}
We supplement more evaluation of multiple benchmarks in Table ~\ref{tab:genimage++},~\ref{tab:evalgen}, and ~\ref{tab:aigi_holmes}.

\paragraph{Detail Setting in Robust Evaluation}
In Table~\ref{tab:degradation}, we exhibit the setting in Robust Evaluation.
The degradation includes Color Contrast adjustment via $P_{out} = \alpha P_{in}$ and Color Saturation modulation via $C_{out} = 0.5 + \beta (C_{in} - 0.5)$, where $\alpha$ and $\beta$ represent the contrast scaling factor and chroma gain coefficient, respectively.
\begin{table*}[h]
\centering
\scriptsize
\caption{Degradation configurations for robustness evaluation.Gaussian blurring was defined by kernel size ($K$) and standard deviation ($\sigma$); Color contrast was adjusted by the linear scaling factor ($\alpha$); Color saturation was modulated by the chroma gain coefficient ($\beta$) in YCbCr space.}
\label{tab:degradation}
\setlength{\tabcolsep}{16pt}
\begin{tabular}{l|ccc}
\toprule
\textbf{Level} & \textbf{Color Contrast} & \textbf{Gaussian Blur} & \textbf{Color Saturation} \\
\midrule
1 & 0.850 & $K=(7,7), \sigma=1$ & 0.4 \\
2 & 0.725 & $K=(9,9), \sigma=2$ & 0.3 \\
3 & 0.600 & $K=(13,13), \sigma=3$ & 0.2 \\
4 & 0.475 & $K=(17,17), \sigma=4$ & 0.1 \\
5 & 0.350 & $K=(21,21), \sigma=5$ & 0.0 \\
\bottomrule
\end{tabular}
\end{table*}

\begin{table*}[t]
\centering
  \scriptsize
\caption{The generalizable performance (ACC (\%)) in GenImage++ Dataset~\citep{zhou2025breaking}. This dataset includes fake images only.} 
\setlength{\tabcolsep}{10pt}
\begin{tabular}{l|ccccccccc|c}
\toprule
Model & Flux & Flux-M & Flux-P & Flux-R & SD1.5-M & SDXL-M & SD3 & SD3-P & SD3-R & \textbf{AVG} \\
\midrule
Xception  & 36.86 & 10.48 & 4.65  & 5.45  & 97.27 & 20.63 & 38.00 & 5.83  & 15.06 & 26.03 \\
CNNSpot   & 37.38 & 6.89  & 8.71  & 5.28  & 84.41 & 34.79 & 47.70 & 7.48  & 25.55 & 28.69 \\
F3Net     & 25.18 & 7.79 & 2.83 & 7.90 & 94.15 & 24.01 & 46.67 & 0.84 & 30.28 & 26.63 \\
GramNet   & 37.83 & 16.71 & 8.01  & 19.71 & 96.49 & 28.65 & 48.55 & 8.33  & 55.71 & 35.55 \\
NPR       & 35.38 & 13.19 & 8.48  & 19.41 & 93.63 & 15.40 & 32.38 & 12.45 & 27.58 & 28.66 \\
SPSL      & 67.13 & 16.55 & 43.76 & 25.73 & 71.14 & 17.74 & 44.58 & 16.22 & 29.75 & 36.96 \\
SRM       & 8.46  & 2.92  & 0.37  & 1.93  & 96.62 & 6.39  & 9.97  & 0.55  & 4.43  & 14.63 \\
DRCT/Conv-B & 73.02 & 51.91 & 54.72 & 66.40 & 100.00 & 77.19 & 79.10 & 82.93 & 76.58 & 73.54\\
DRCT/UniFD & 71.08 & 63.97 & 46.83 & 62.42 & 99.19 & 64.84 & 72.28 & 70.70 & 73.55 & 69.43 \\
OMAT & {96.53} & {92.55} & {97.60} & {97.67} & \textbf{100.00} & \textbf{99.17} &{98.27} & {90.38} & {98.82} & {96.78} \\
\midrule
\rowcolor{blue!3} \textbf{Deep-VRM} &\textbf{99.85} &\textbf{99.95} &\textbf{100.00} &\textbf{100.00} &97.33 &{98.98} &\textbf{99.20} &\textbf{100.00} &{99.55} &\textbf{99.43} \\
\bottomrule
\end{tabular}
\label{tab:genimage++}
\end{table*}

\begin{table*}
  \centering
  \caption{Generalizable performance (ACC (\%)) on EvalGen~\citep{chen2025dual}. We obtained the experimental results from the original paper. This dataset includes fake images only.}
  \label{tab:evalgen}
  \footnotesize
  \setlength{\tabcolsep}{18pt}
  \begin{tabular}{l|ccccccccc}
    \toprule
    Method &Flux &GoT &Infinity &OmiGen &NOVA &\textbf{AVG} \\
    \midrule
    UnivFD &4.00 &9.20 &15.70 &8.30 &39.60 &15.40 \\
    FatFormer &9.90 &47.80 &44.70 &98.30 &27.30 &45.60 \\
    C2P-CLIP &8.70 &49.40 &35.10 &86.40 &14.80 &38.90 \\
    AIDE &16.20 &21.60 &4.00 &14.90 &18.40 &15.00 \\
    AlignedForensics &45.00 &84.40 &79.60 &90.80 &85.20 &77.00 \\
    DDA &87.00 &99.30 &99.50 &{99.50} &\textbf{100.00} &94.00 \\
    \midrule

    \rowcolor{blue!3}  \textbf{Deep-VRM} &\textbf{100.00} &\textbf{100.00} &\textbf{100.00} &\textbf{100.00} &\textbf{100.00} &\textbf{100.00} \\
    
    \bottomrule
  \end{tabular}
\end{table*}

\begin{table*}
  \centering
  \caption{Performance (ACC (\%)) on evaluation set in Community Forensics~\citep{park2025community}. We ignored the subset `DALLE2' in this table.}
  \label{tab: comfor}
  \resizebox{\textwidth}{!}{
  \begin{tabular}{l|cccccccccccc}
    \toprule
    Method &DALLE3 &DFGAN &Flux-dev &GALIP &Hourglass &IdeogramV1 &IdeogramV2 &Imagen3   \\
    \midrule

    C2P-CLIP &64.15 &99.30 &60.25 &74.44 &68.70 &51.45 &50.80 &50.38 \\
    DeeCLIP &91.00 &{99.85} &57.70 &\textbf{86.35} &65.75 &67.55 &62.15 &78.43 \\
    DRCT &94.80 &50.65 &88.70 &53.50 &52.40 &92.05 &90.80 &93.38 \\
    PatchShuffle &\textbf{99.85} &66.15 &98.25 &51.70 &65.65 &96.90 &96.20 &98.85 \\
    \rowcolor{blue!3}  \textbf{Deep-VRM} &99.80 &\textbf{99.95} &\textbf{99.90} &81.75 &96.55 &\textbf{99.80} &\textbf{99.35} &\textbf{99.91} \\
    \midrule
    \midrule
    Method &Kandinsky &Kvikontent &LCM-SD15 &LCM-SDXL &LCM-SSD1B &MidJourney V5 &Stable Cascade &\textbf{AVG} \\
    \midrule
    C2P-CLIP &58.00 &85.05 &82.30 &54.05 &89.10 &54.14 &65.60 &67.18 \\
    DeeCLIP &80.15 &97.85 &55.95 &55.65 &56.85 &75.97 &94.60 &75.05 \\
    DRCT &{99.75} &98.65 &95.85 &\textbf{97.90} &89.85 &{97.82} &97.90 &86.27 \\
    PatchShuffle &99.50 &99.30 &92.55 &71.30  &58.70 &81.51 &94.40 &84.72 \\
    \rowcolor{blue!3}  \textbf{Deep-VRM} &\textbf{99.95} &\textbf{99.70} &99.55 &96.30 &\textbf{96.10} &\textbf{99.97} &\textbf{99.90} &\textbf{97.90} \\
    \bottomrule
  \end{tabular}
}
\end{table*}

\begin{table*}[t]
\centering

\caption{Generalization performance (ACC \%) on the AIGI-Holmes dataset~\citep{zhou2025aigi}. Baseline results are cited from the original paper. Note that AIGI-Holmes$^{*}$ denotes the standalone MLLM setting, excluding the ensemble with external dedicated detectors.} 
\resizebox{\linewidth}{!}{
\begin{tabular}{l|cccccccccc|c}
\toprule
Model & Janus & J-Pro-1B &J-Pro-7B & Show-o & LlamaGen & Infinity & VAR & PixArt-XL &SD3.5 L & FLUX & \textbf{AVG} \\
\midrule
CNNSpot &70.00 &70.90 &85.00 &72.20 &61.90 &86.80 &59.90 &78.20 &63.80 &79.90 &72.90 \\
AntiFakePrompt &72.20 &84.30 &84.80 &86.20 &96.20 &83.60 &90.70 &81.70 &92.80 &66.10 &83.90 \\
UnivFD &87.60 &96.90 &96.40 &85.90 &93.10 &79.20 &64.30 &75.70 &87.80 &69.60 &83.60 \\
NPR &51.20 &69.50 &73.90 &93.70 &93.50 &93.80 &85.90 &93.40 &91.60 &93.60 &84.00 \\
LaRE &70.80 &74.70 &95.60 &80.00 &91.60 &77.90 &\textbf{98.80} &82.20 &94.10 &84.30 &85.00 \\
RINE &89.90 &98.70 &97.20 &98.80 &99.10 &99.20 &85.00 &98.90 &97.80 &97.10 &96.20 \\
AIDE &\textbf{91.20} &\textbf{98.90} &\textbf{97.80} &98.00 &99.40 &98.70 &93.60 &98.60 &{99.40} &94.40 &\textbf{97.00} \\
\midrule
AIGI-Holmes$^{*}$ &80.20 &91.90 &89.60 &98.00 &98.00 &98.40 &76.00 &98.50 &97.80 &94.20 & 92.30 \\
\midrule
\rowcolor{blue!3} \textbf{Deep-VRM} &{77.40} &96.12 &96.98 &\textbf{99.95} &\textbf{99.98} &\textbf{99.95} &{91.45} &\textbf{99.93} &\textbf{99.62} &\textbf{99.67} &{96.11} \\

\bottomrule
\end{tabular}
}
\label{tab:aigi_holmes}
\end{table*}

\end{document}